\documentclass[graybox]{styles/svmult}

\usepackage{mathptmx}       
\usepackage{helvet}         
\usepackage{courier}        
\usepackage{type1cm}        
\usepackage{makeidx}         
\usepackage{graphicx}        
\usepackage{multicol}        
\usepackage{multirow}
\usepackage{mdframed}
\usepackage[bottom]{footmisc}
\usepackage{color}

\makeindex             

\begin{document}

\title*{Response Generation for Cognitive Behavioral Therapy with Large Language Models: Comparative Study with Socratic Questioning}
\titlerunning{Response Generation for CBT with LLMs}
\author{Kenta Izumi, Hiroki Tanaka, Kazuhiro Shidara, Hiroyoshi Adachi, Daisuke Kanayama, Takashi Kudo, and Satoshi Nakamura}
\authorrunning{Kenta Izumi et al.}
\institute{Kenta Izumi, Hiroki Tanaka, Kazuhiro Shidara and Satoshi Nakamura \at Nara Institute of Science and Technology, Takayama-cho 8916-5, Ikoma-shi, Nara, Japan. \\ \email{\{izumi.kenta.if6, hiroki-tan, shidara.kazuhiro.sc5, s-nakamura \}@is.naist.jp}
\and Hiroyoshi Adachi, Daisuke Kanayama, and Takashi Kudo \at Health and Counseling Center, Osaka University, Yamadaoka 1-1, Suita-shi, Osaka, Japan.
}

\maketitle

\abstract{
Dialogue systems controlled by predefined or rule-based scenarios derived from counseling techniques, such as cognitive behavioral therapy (CBT), play an important role in mental health apps. 
Despite the need for responsible responses, it is conceivable that using the newly emerging LLMs to generate contextually relevant utterances will enhance these apps. In this study, we construct dialogue modules based on a CBT scenario focused on conventional Socratic questioning using two kinds of LLMs: a Transformer-based dialogue model further trained with a social media empathetic counseling dataset, provided by Osaka Prefecture (OsakaED), and GPT-4, a state-of-the art LLM created by OpenAI. 
By comparing systems that use LLM-generated responses with those that do not, we investigate the impact of generated responses on subjective evaluations such as mood change, cognitive change, and dialogue quality (e.g., empathy). 
As a result, no notable improvements are observed when using the OsakaED model. When using GPT-4, the amount of mood change, empathy, and other dialogue qualities improve significantly. 
Results suggest that GPT-4 possesses a high counseling ability. 
However, they also indicate that even when using a dialogue model trained with a human counseling dataset, it does not necessarily yield better outcomes compared to scenario-based dialogues.
While presenting LLM-generated responses, including GPT-4, and having them interact directly with users in real-life mental health care services may raise ethical issues, it is still possible for human professionals to produce example responses or response templates using LLMs in advance in systems that use rules, scenarios, or example responses.
}

\section{Introduction}
\label{sec:1}

The number of people with mental disorders worldwide is estimated to be around 970 million in 2019, having increased by 25\% since 2000 with the growth of the world population \cite{GBDmentaldisorder}. 
In particular, anxiety disorders and depression accounted for approximately 60\% of the prevalence of mental disorders in 2019 \cite{GBDprevalence}. 
Under these circumstances, qualified psychiatrists and other therapists remain in short supply, and thus the growing number of patients who are unable to receive the care they need has become a major concern \cite{ShortageOfPsychiatry, BrainDrain}. 
On the other hand, with the spread of smartphones, various mobile health apps have been developed, and it's expected that they will help to provide immediate care to people who have difficulty accessing needed healthcare \cite{mHealthGap, mHealthGrowth}. 
Among other solutions, dialogue systems such as Wysa \cite{inkster2018empathy}, which encourages users to build cognitive resilience through empathetic responses based on counseling techniques such as cognitive behavioral therapy (CBT) \cite{beck1979cognitive}, are expected to be useful in the form of mental healthcare apps. 
Toward this approach, a variety of dialogue systems have been explored so far \cite{vaidyam2019chatbots,abd2019overview,shidara2022automatic, akiyoshi2021robot}. 

A review of the literature up to 2019 reported that most of these systems generate responses based on pre-defined rules, and thus system utterances are commonly described as inadequate, e.g., ``conversations are shallow, confusing or too short." Accordingly, the adoption of natural language processing technology is expected to improve the quality of dialogue \cite{abd2021perceptions}. 
Recently, natural language understanding and generation technology for large language models (LLMs), as exemplified by the 2022 release of OpenAI's ChatGPT \cite{chatGPT}, has developed rapidly, and CBT dialogue systems that use high-performance LLMs to generate responses have also been researched and developed  \cite{danieli2022assessing,liu2023chatcounselor}.
Most previous apps had been controlled by predefined or rule-based scenarios,
mainly due to technological limitations. 
Nevertheless, they had the advantage of providing responsible responses, compared to the LLM-generated responses that the developers cannot fully control and predict.

To the best of our knowledge, no study has examined how the use of LLM-based response generation changes the effectiveness of CBT and user experiences compared to conventionally adopted simple scenario or rule-based systems. 
We examined whether the user's subjective evaluation of the system was improved by replacing or combining the Socratic questions in a scenario with responses generated by LLMs.

\section{Methods}
\label{sec:2}

In this study, dialogue scenarios were created with reference to the CBT dialogue scenarios centered on Socratic questions, considered important in CBT, as used in Shidara et al. \cite{shidara2022automatic}. 
We conducted an experiment to investigate the effect on the user's mood and cognitive change \cite{persons1985mechanisms,schmidt2019putting}, which are indices for measuring the effectiveness of CBT, and other dialogue qualities when the Socratic questions in a scenario were replaced or combined with responses generated by an LLM.

\subsection{Dialogue scenarios}
\label{subsec:2}


In this study, scenarios consisting of 15 system utterances were created for the CBT-based dialogue scenario \cite{shidara2022automatic} (Table \ref{tab:sce}). 
In CBT, when a certain mood arises, the thoughts that occur automatically, such as assumptions and interpretations of the situation, are called automatic thoughts. Patients are encouraged to modify these into more balanced and adaptive thoughts \cite{beck2020cognitive}. 
In the scenario, the system first asks the user about the situation that is troubling him/her, then asks about the mood that arises when she/he thinks about it and its intensity, and finally asks about the thoughts that occur at that time and puts the strongest thoughts on the agenda as automatic thoughts (utterance numbers 1-7). 
This part can be considered questions asking about ABC: A) Activating event, B) Beliefs, and C) Consequences, which is a basic model in CBT \cite{ellis1991revised}.
Then, six turns of Socratic questions are asked to encourage the user to objectively view and reconsider his/her automatic thoughts (utterance numbers 8-13). 
In addition, the user is again asked about the intensity of her/his current mood and the scenario ends with a parting expression (utterance numbers 14 and 15). 

Socratic questioning is a questioning technique that allows the therapist to explore the client's thoughts and beliefs and to help the client think critically about them, broaden their perspective, and arrive at a more realistic and adaptive perspective. 
This technique has been considered an important component of CBT for depression \cite{vittorio2022using}.

\begin{table}[h]
\caption{Dialogue scenario for the dialogue system in SQ condition. Question items and question order are the same in all experiments. Only questions in bold differ from the LLM condition; the conditions with LLMs use only generated sentences or combinations of a Socratic question and a generated sentence.}
\label{tab:sce}       
\scalebox{0.88}{
\begin{tabular}{ll|l}
\hline
\multirow{2}{*}{}            & \multirow{2}{*}{Question item}                                                          & \multirow{2}{*}{System's response}                                                                                                                                                                                                                                                         \\
                             &                                                                                         &                                                                                                                                                                                                                                                                                            \\ \hline
\multirow{2}{*}{1}           & \multirow{2}{*}{Situation}                                                              & \multirow{2}{*}{\begin{tabular}[c]{@{}l@{}}Let's address your concerns together. Are you currently facing any issues or\\ difficulties?  If you feel comfortable doing so, please share what's troubling you.\end{tabular}}                                                              \\
                             &                                                                                         &                                                                                                                                                                                                                                                                                            \\ \hline
\multirow{2}{*}{2}           & \multirow{2}{*}{Mood}                                                                   & \multirow{2}{*}{How did you feel at that time?}                                                                                                                                                                                                                                            \\
                             &                                                                                         &                                                                                                                                                                                                                                                                                            \\ \hline
\multirow{2}{*}{3}           & \multirow{2}{*}{\begin{tabular}[c]{@{}l@{}}Initial mood\\score\end{tabular}}  & \multirow{2}{*}{\begin{tabular}[c]{@{}l@{}}On a scale of 0 to 100, where 0 is no problem at all and 100 is   \\ a huge problem, how intense is that feeling?\end{tabular}}                                                                                                                 \\
                             &                                                                                         &                                                                                                                                                                                                                                                                                            \\ \hline
\multirow{2}{*}{4}           & \multirow{2}{*}{\begin{tabular}[c]{@{}l@{}}Automatic\\thought\end{tabular}}            & \multirow{2}{*}{What thoughts came to your mind at that time?}                                                                                                                                                                                                                             \\
                             &                                                                                         &                                                                                                                                                                                                                                                                                            \\ \hline
\multirow{2}{*}{5}           & \multirow{2}{*}{\begin{tabular}[c]{@{}l@{}}Automatic\\thought\end{tabular}}            & \multirow{2}{*}{I see. Are there any others? Please   share anything that comes to mind.}                                                                                                                                                                                                  \\
                             &                                                                                         &                                                                                                                                                                                                                                                                                            \\ \hline
\multirow{2}{*}{6}           & \multirow{2}{*}{\begin{tabular}[c]{@{}l@{}}Automatic\\thought\end{tabular}}            & \multirow{2}{*}{Please tell me the strongest thought   among those that came to your mind.}                                                                                                                                                                                                \\
                             &                                                                                         &                                                                                                                                                                                                                                                                                            \\ \hline
\multirow{2}{*}{7}           & \multirow{2}{*}{\begin{tabular}[c]{@{}l@{}}Automatic\\thought\end{tabular}}            & \multirow{2}{*}{\begin{tabular}[c]{@{}l@{}}The thoughts that come to mind in response to a situation are called automatic thoughts.\\ Today we are going to delve into your thought. Is that okay?\end{tabular}} \\
                             &                                                                                         &                                                                                                                                                                                                                                                                                            \\ \hline
\multirow{2}{*}{\textbf{8}}  & \multirow{2}{*}{\textbf{\begin{tabular}[c]{@{}l@{}}Socratic\\ question 1\end{tabular}}} & \multirow{2}{*}{If the automatic thought is correct, what is the basis for it?}                                                                                                                                                                                                          \\
                             &                                                                                         &                                                                                                                                                                                                                                                                                            \\ \hline
\multirow{2}{*}{\textbf{9}}  & \multirow{2}{*}{\textbf{\begin{tabular}[c]{@{}l@{}}Socratic\\ question 2\end{tabular}}} & \multirow{2}{*}{On the other hand, if your automatic thought is incorrect, what evidence can   you think of?}                                                                                                                                                                             \\
                             &                                                                                         &                                                                                                                                                                                                                                                                                            \\ \hline
\multirow{2}{*}{\textbf{10}} & \multirow{2}{*}{\textbf{\begin{tabular}[c]{@{}l@{}}Socratic\\ question 3\end{tabular}}} & \multirow{2}{*}{\begin{tabular}[c]{@{}l@{}}Can you think of any other perspectives regarding the situation that's troubling you, \\ aside from your current automatic thoughts?\end{tabular}}                                                                                            \\
                             &                                                                                         &                                                                                                                                                                                                                                                                                            \\ \hline
\multirow{2}{*}{\textbf{11}} & \multirow{2}{*}{\textbf{\begin{tabular}[c]{@{}l@{}}Socratic\\ question 4\end{tabular}}} & \multirow{2}{*}{What would be the worst possible outcome in this situation?}                                                                                                                                                                                                               \\
                             &                                                                                         &                                                                                                                                                                                                                                                                                            \\ \hline
\multirow{2}{*}{\textbf{12}} & \multirow{2}{*}{\textbf{\begin{tabular}[c]{@{}l@{}}Socratic\\ question 5\end{tabular}}} & \multirow{2}{*}{What would be the best possible outcome in this situation?}                                                                                                                                                                                                                \\
                             &                                                                                         &                                                                                                                                                                                                                                                                                            \\ \hline
\multirow{2}{*}{\textbf{13}} & \multirow{2}{*}{\textbf{\begin{tabular}[c]{@{}l@{}}Socratic\\ question 6\end{tabular}}} & \multirow{2}{*}{What do you think is the most realistic outcome in this situation?}                                                                                                                                                                                                        \\
                             &                                                                                         &                                                                                                                                                                                                                                                                                            \\ \hline
\multirow{2}{*}{14}          & \multirow{2}{*}{\begin{tabular}[c]{@{}l@{}}Mood score \\ at the end\end{tabular}}       & \multirow{2}{*}{\begin{tabular}[c]{@{}l@{}}Now, how intense is the feeling you initially had? \\ Please rate it again on a scale from 0 to 100.\end{tabular}}                                                                                                                              \\
                             &                                                                                         &                                                                                                                                                                                                                                                                                            \\ \hline
\multirow{2}{*}{15}          & \multirow{2}{*}{\begin{tabular}[c]{@{}l@{}}Closing \\ remark\end{tabular}}                                                        & \multirow{2}{*}{\begin{tabular}[c]{@{}l@{}}If your mood has changed, that’s a sign that you've successfully organized your thoughts. \\That's it for today. Good job!\end{tabular}}                                                                                                       \\
                             &                                                                                         &                                                                                                                                                                                                                                                                                            \\ \hline
\end{tabular}
}
\end{table}

In this study, we implemented the following three types of dialogue systems, where utterances 8-13 of the scenarios were varied according to the system: 1) asking Socratic questions as in the scenario, 2) uttering the responses generated by the LLMs, and 3) combining LLM-generated responses with Socratic questions.
We focus on how each dialogue system can help people rethink their automatic thought, arrive at a more adaptive and balanced thought, and thus improve their mood after asking ABC questions (in utterance numbers 1-7).
In particular, the dialogue flow of system 2, where the system asks ABC questions and utters generated responses after that, is roughly similar to TEO (Therapy Empowerment Opportunity) used in Danieli et al. \cite{danieli2022assessing, mousavi2021Wouldyoulike}.
We expected that adding Socratic questions to LLM-generated responses would help the systems encourage the user to rethink their automatic thought and not deviate from the context of CBT.
In each of the latter two systems, which generate responses by LLMs, the following two LLMs were used to implement the dialogue systems: Osaka empathetic counseling dialogue model (OsakaED) and GPT-4.
We chose these two LLMs because they have different characteristics from each other: OsakaED specializes in counseling in social media, and GPT-4 is a general-purpose state-of-the-art model.
In this way, five dialogue systems were implemented, each with different utterances 8-13 of the dialogue scenario.
Socratic questions are represented by SQ and OsakaED, and when Socratic questions are added to the utterances of each model, they are connected by ``+" to form the system name. 

\begin{table}
\caption{List of dialogue systems compared in this study.}
\label{tab:1}       

\begin{tabular}{p{2cm}p{9.5cm}p{2cm}p{4.9cm}}
\hline\noalign{\smallskip}
System name & Description  \\
\noalign{\smallskip}\svhline\noalign{\smallskip}
SQ & Socratic questions based on the scenario are uttered\\
OsakaED & Response generated by Osaka social media empathetic dialogue model are uttered\\
OsakaED+SQ & Socratic questions based on the scenario are added to the responses generated by Osaka social media empathetic dialogue model and uttered\\
GPT-4 & Response generated by GPT-4 are uttered\\
GPT-4+SQ & Socratic questions based on the scenario are added to the responses generated by GPT-4 and uttered\\
\noalign{\smallskip}\hline\noalign{\smallskip}
\end{tabular}

\end{table}

\paragraph{OsakaED}
OsakaED was developed by further fine-tuning the Japanese Transformer Encoder-Decoder dialogue model with a dataset provided by Osaka Prefecture. 
This model, created by Sugiyama et al. \cite{sugiyama2023empirical}, had already been fine-tuned with empathic dialogue data and possesses 1.6 billion parameters.
We chose this model as a base model because it's trained with text chat conversations via the internet and thought to already have empathetic response generation ability in a form similar to human speech on social media.
Osaka social media counseling dataset consists of 2077 dialogues and 62,066 utterances of counselors counseling their clients using the messenger application LINE\footnote{https://line.me/}. 
From these, 29,062 pairs of context and counselor utterances were extracted so as not to include utterances consisting only of system utterances, which are mainly placed at the beginning of the dialogues.
In training, 90\% of the data were training data and 10\% were validation data, with a batch size of 4, cross-entropy as the objective function, and Adafactor as the optimizer. Here, 11 epochs were trained with a learning rate of $1.0 \times 10^{-5}$, and the validation perplexity used the model from the 6th epoch with the lowest validation perplexity.
Because of the limitation of the input length, the 1 to 7 user utterances are combined and are treated as one utterance, which represents an ABC note, and the system utterance 8 was generated to follow it.
System utterances 9 to 13 were generated by entering the ABC notes and subsequent system and user utterances as history.
The input preprocessing and response generation method was basically the same as Sugiyama's implementation \cite{sugiyama2021empirical}, which sampled five response candidates by top-p sampling with $p=0.90$ and ranked them.
In the dialogues in the pre-experiment, keywords that were included in the generated responses, which attempted to end or guide the dialogues specific to the dataset and were considered inappropriate, were observed and set as NG (No Good) words.
The highest rank response not containing any of NG words was selected as the response. 
When NG words are included in all response candidates, an error message is output, and in this case, the sample is not treated as valid in the subjective evaluation.
In addition, when Socratic questions were added, the response with the highest rank not containing any of NG words and not ending with a question was selected to avoid duplicating questions.
If all response candidates were questions, the response with the highest rank among the candidates that did not contain any of the NG words was selected as the response.


\paragraph{GPT-4}
GPT-4 is a large-scale language model developed by OpenAI that shows human-level performance on tasks across a wide range of domains, including mathematics, coding, and medicine \cite{bubeck2023sparks}.
We used GPT-4 through OpenAI's Chat Completions API.
The Chat Completions API pairs a role ``role" with its message content ``content" in a dictionary format, and then compiles these multiple dictionaries into a list in the order of utterances, which is input as ``messages".
If ``user'' is specified as the ``role", the ``content'' is the content of the message from the user (user message), and if ``assistant'' is specified, it is the speech generated by GPT-4 (assistant message).
when ``system'' is specified, it describes how the assistant works in the whole conversation (system message), such as instructions to the assistant and his/her personality.
Usually, the system message is placed at the beginning of the history, followed by the user message and the assistant message alternating \cite{openaiapi}.
In this study, we followed this method to input messages.

Kumar et al. explored prompt design for mental health chatbots using GPT-3, focusing on the three dimensions of identity, intent, and behavior, with two identities and three intents and behaviors, respectively, for a total of 18 different dialogue systems, and thus 18 prompts were created and evaluated.
The prompts with ``coach" as the identity and ``strong interpersonal skills" as the behavior received the highest overall evaluation among the prompts with intent for CBT \cite{kumar2022exploring}, and it was adopted as the system message.
In this study, the system messages shown in Table \ref{SystemPrompt}, described below, were placed first, and the dialogue history, including subsequent user and system utterances, was included in the messages as much as possible to generate responses as input to GPT-4.
Temperature was set to 1 the same as the default setting.
When adding a Socratic question to the generated response, as shown in Table \ref{SystemPrompt}, we have ``the coach does not ask questions." This sentence was inserted after the first sentence to discourage the respondent from duplicating the question.

\begin{table*}[]
    \centering
    \caption{System prompts for GPT-4 and GPT-4+SQ, based on Kumar et al. \cite{kumar2022exploring}. The prompts translated into Japanese without indentation were used in the experiment.}
    \label{SystemPrompt}
    \begin{tabular}{c|l} \hline
        System  & System prompt\\ \hline
        GPT-4 & 
        \begin{tabular}{l}
        The following is a conversation with a coach.\\
        The coach helps the Human understand how their thoughts, feelings, \\
        and behaviors influence each other. \\
        If the Human demonstrates negative thoughts, \\
        the coach helps the Human replace them with more realistic beliefs.\\
        The coach has strong interpersonal skills.
        \end{tabular}\\ \hline
        GPT-4+SQ & 
        \begin{tabular}{l}
        The following is a conversation with a coach.\\
        \textbf{The coach does not ask questions.}\\
        The coach helps the Human understand how their thoughts, feelings, \\
        and behaviors influence each other. \\
        If the Human demonstrates negative thoughts, \\
        the coach helps the Human replace them with more realistic beliefs.\\
        The coach has strong interpersonal skills.
        \end{tabular}\\ \hline 
    \end{tabular}
    
\end{table*}

Example responses from GPT-4 and OsakaED are shown below.
Both examples consist of the system utterances 8 and 9 and the user utterance 8, in the context when the user consults about skin trouble and acne.

\begin{mdframed}
\begin{description}[GPT-4 example] 
\item[GPT-4]{Thank you. Let's dig deeper into the thoughts that arise from that ``I want to heal fast" feeling. How do you feel that thought is affecting you? How does it affect your behavior and daily life?}
\item[User]{I check when looking in the mirror and pay attention to the skin and diet.}
\item[GPT-4]{Your automatic thoughts strongly express your desire to ``heal quickly". This thought prompts you to take action and make it easier for you to try to live a healthier life. You once thought, ``Why can't I get well?" but in fact, that thought may not be in alignment with reality. I would like to propose a new way of looking at it. You can help yourself more by looking at it from a realistic perspective, as in, ``I have control over my skin condition," and ``I can improve it over time if I take proper care of my skin problems." What do you think about this new view? }
\end{description} 
\end{mdframed}

\begin{mdframed}
\begin{description}[Osaka example]
\item[OsakaED]{It's hard to have skin problems and acne that won't go away for a long time...}
\item[User]{Yes, it is. It's hard.}
\item[OsakaED]{Why is it that skin problems and acne don't go away all the time...}
\end{description}
\end{mdframed}

As shown in these examples, GPT-4 tends to help the user explore their thoughts in a longer text, while OsakaED show empathy in a shorter text.

\subsection{Subjective evaluation}
A subjective evaluation experiment was conducted using a crowdsourcing platform, CrowdWorks \footnote{https://crowdworks.jp/}, in which 105 crowd participants (healthy participants) were gathered to evaluate our dialogue systems.
This experiment was conducted under the review and approval of the research ethics committee of the Nara Institute of Science and Technology (reference number: 2022-I-52).
After reading the explanation of the experiment and agreeing to participate in the experiment, the participants were asked to provide their crowdsourcing ID, gender, age, and rating on the Kessler Psychological Distress Scale (K6) \cite{furukawa2003performance}, a widely used measure of psychological distress, to be assigned to each participant at CrowdWorks.
They then read a leaflet about CBT written by a psychiatrist\footnote{https://www.cbtjp.net/downloads/skillup/}.
The participants reserved a time to perform a dialogue with the system, started the dialogue at the reserved time, and ended the dialogue after confirming the last utterance of the system (utterance number 15).
The participants completed the experiment by answering a questionnaire about cognitive change and dialogue quality, and free-text descriptions about the strengths, weaknesses, and improvements of the system.
Throughout the experiment, participants were not informed of the type of dialogue system with which they interacted.


In CBT, the goal is to improve mood by modifying automatic thoughts into adaptive thoughts.
In this experiment, following the work of Persons et al. \cite{persons1985mechanisms}, the mood change was calculated by an equation \ref{eq1} and then used as an index to evaluate the degree to which mood improved.
Here, $initial\, mood\, score$ and $mood\, score\, at\, end$ represent the intensities of mood elicited by system utterances 3 and 14 in the scenario, respectively: 

\begin{equation}
\label{eq1}
    Mood\, change 
    = -\frac{\left(mood\, score\, at\, end \right) - \left(initial\, mood\, score \right)}{\left(initial\, mood\, score \right)}
\end{equation}

The Cognitive Change immediate scale (CC-immediate) \cite{schmidt2019putting} for measuring immediate cognitive change with single-session CBT was translated into Japanese and used to assess cognitive change due to interaction with the system.
The CC-immediate consists of six questions, each rated on a 7-point Likert scale with 0 being ``not at all" and 6 being ``completely," and the responses to all questions were summed to form a 31-point scale from 0 to 30 to evaluate the immediate change in cognition. 

Additionally, a questionnaire consisting of 15 items was used to evaluate the quality of dialogue at the end of the conversation, which has two additional questions to that created by Sugiyama et al. \cite{sugiyama2023empirical}, based on a survey for dialogue between dialogue systems and users conducted by Fitrianie et al. \cite{fitrianie2020}. 
The Sugiyama et al. study originally included an item called ``empathy," but because this item asked whether the user could empathize with the system's utterances, ``I could empathize with the system's utterances," it was changed to an item called user's empathy, and a new item called ``system's empathy" was added along with the question ``the system showed empathy" \cite{sugiyama2023empirical}.
This is because assessing the system's empathetic ability has been suggested as an effective way of soothing the user's mood and, more specifically, improving the outcome \cite{de2020effectiveness,jani012roleofempathy}.
The item called ``initiative" with the question ``I was able to lead the dialogue" was also added.
It's pointed out to be lost in scenarios and rule-based dialogues in a study by Abd-Arazaq et al. \cite{abd2021perceptions}, and we assess this to confirm whether this quality is improved by using LLMs and its importance for CBT.
Continued usability is measured by the item ``re-speak," which is measured by the question ``I would like to talk to this system again." 
Each item was rated on an 11-point Likert scale, with 0 representing ``completely disagree" and 10 representing ``completely agree." 
Mann-Whitney's U-test was used to assess the difference in scales between the two systems according to the particular interests of comparison.

\section{Results}
\label{sec:3}

The experiment resulted in a total of 86 valid responses. The number of users for each system was 24 for SQ, 14 for OsakaED, 13 for OsakaED+SQ, 17 for GPT-4, and 18 for GPT-4+SQ.
The mean and standard deviation of K6 was 7.60 and 5.59, and no significant differences were seen in any two groups' comparisons ($p>0.27$ in all comparisons).
The results of subjective evaluations are summarized in Table \ref{Result}.

\begin{table}
\caption{Subjective evaluations of each system (mean and standard deviation). Bold font represents the highest value in each scale.}
\label{Result}       
%

\scalebox{0.9}{
\begin{tabular}{p{2.2cm}p{2cm}p{2cm}p{2cm}p{2cm}p{2cm}}
\hline\noalign{\smallskip}
 & SQ & OsakaED & OsakaED+SQ & GPT-4 & GPT-4+SQ  \\
\noalign{\smallskip}\svhline\noalign{\smallskip}

mood change     & 0.28 (0.24) & 0.22 (0.16) & 0.19 (0.40) & 0.44 (0.26) & \textbf{0.49 (0.14)} \\
CC-immediate    & 16.6 (7.07) & 14.4 (5.83) & 16.9 (4.65) & \textbf{19.8 (4.93)} & 18.8 (5.55) \\ \hline
naturalness     & 6.25 (2.97) & 5.93 (2.56) & 6.54 (1.61) & \textbf{7.76 (2.19)} & 6.56 (2.89) \\
ease            & 7.25 (2.45) & 7.00 (2.94) & \textbf{8.31 (1.55)} & 8.06 (1.43) & 6.83 (3.13) \\
enjoyment       & 5.13 (2.98) & 5.36 (2.92) & 6.23 (2.24) & \textbf{6.53 (2.70)} & 6.39 (2.68) \\
user's empathy  & 5.17 (2.62) & 5.07 (2.62) & 5.85 (1.72) & \textbf{7.76 (2.14)} & 7.06 (2.10) \\
attentiveness   & 5.79 (2.95) & 6.14 (3.01) & 7.15 (2.15) & \textbf{7.59 (2.29)} & 7.56 (1.46) \\
trust           & 5.63 (2.65) & 5.43 (2.90) & 5.77 (1.92) & \textbf{7.76 (2.19)} & 7.44 (1.82) \\
personality     & 3.96 (3.04) & 4.79 (2.91) & 5.00 (2.45) & \textbf{6.65 (2.94)} & 6.33 (2.33) \\
agency          & 4.29 (3.09) & 4.57 (2.82) & 4.85 (2.64) & \textbf{7.18 (2.38)} & 6.22 (2.34) \\
topic           & 5.54 (3.20) & 4.21 (2.12) & 5.54 (2.60) & \textbf{5.94 (2.84)} & 5.72 (2.49) \\
emotion         & 3.42 (2.93) & 4.29 (2.52) & 4.00 (2.38) & \textbf{6.41 (2.60)} & 5.50 (2.71) \\
consistency     & 8.04 (1.68) & 5.79 (2.83) & 6.77 (1.59) & \textbf{8.47 (1.42)} & 8.00 (2.00) \\
engagement      & 5.63 (3.21) & 5.43 (2.56) & 5.46 (2.50) & \textbf{7.06 (2.44)} & 6.39 (2.33) \\
re-speak         & 5.50 (3.12) & 5.93 (2.81) & 6.15 (2.44) & \textbf{6.65 (2.74)} & 6.50 (2.98) \\
system's empathy& 5.42 (2.62) & 6.93 (2.30) & 6.85 (2.70) & \textbf{7.94 (2.14)} & 7.44 (2.31) \\
initiative      & 4.21 (2.32) & \textbf{5.79 (1.85)} & 5.46 (2.26) & 5.53 (2.60) & 4.94 (2.96) \\
\noalign{\smallskip}\hline\noalign{\smallskip}
\end{tabular}
}
\end{table}

An overview of Table 4 shows that GPT-4 received the highest ratings in 14 items except for mood change, ease, and initiative.
For mood change, ease, and initiative, GPT-4+SQ, OsakaED+SQ, and OsakaED received the highest ratings, respectively.
The two systems using GPT-4 received higher ratings than the other systems for user's emapthy, trust, personality, agency, emotion, engagement, and system's empathy, as well as for mood change, and CC-immediate.
These more emotionally rich and empathetic responses may have led to improved outcomes such as mood change and CC-immediate.
On the other hand, the two systems using the OsakaED did not score as highly on many items as SQ, although they did score highly on ease and initiative.
The mood change and CC-immediate did not show improvements compared to SQ, and it is not considered that the outcome was improved.


In the following, we will analyze the results in detail by focusing on specific points of view and comparing two groups.

\subsection{Comparison between SQ and generated responses}

To investigate the effect of replacing SQ with generated responses from LLMs, the differences between SQ and OsakaED and between SQ and GPT-4 were assessed.
In the comparison between SQ and OsakaED, there were significant differences in consistency and initiative: Consistency decreased ($p<0.05$), while initiative increased ($p<0.05$).
In the comparison between SQ and GPT-4, there were significant differences in mood change and attentiveness ($p<0.05$), as well as in user empathy, trust, personality, agency, emotion, and the system's empathy ($p<0.01$). 

In scenarios where SQ was replaced with generated responses from OsakaED, there was only one improvement in initiative. 
Considering that OsakaED performed better in many of the items, including consistency and user's empathy, than with the reported result of the base model (model trained with EmpatheticDialogue dataset) in the previous work \cite{sugiyama2023empirical}, it's surprising that simply asking Socratic questions in six turns worked competitively.
The increase in initiative suggests that LLM responses may prompt users to take more control or be more proactive in the interaction.
Conversely, the decrease in consistency indicates that OsakaED-generated responses may be more variable or less predictable than SQ. 

The significant differences observed when SQ was replaced with GPT-4 responses in terms of mood change and attentiveness, as well as a range of user experience dimensions, emphasize the sophisticated capabilities of GPT-4 in affecting user perception, even though the provided prompt is not optimized for this very well. 
The positive changes in mood, user and system empathy, and emotion are particularly noteworthy, since they suggest that interactions with GPT-4 can lead to an emotionally satisfying user experience. 
However, in many of these improvements, re-speak is not significantly improved ($p=0.23$).
For long-time use, this dimension is believed to be important, and thus further consideration should be given to it.
Additionally, despite the prompt being designed to discourage duplication of questioning for GPT-4+SQ, it was observed that responses generated by both GPT-4 and GPT-4+SQ frequently involved Socratic Questions.
An annotator assigned to analyze these responses reported that they consistently involved Socratic Questions in all dialogues.
This indicates that GPT-4 not only understands Socratic Questions but also has the capability to appropriately ask them in context.

\subsection{Comparison between SQ and generated responses with SQ}

To investigate the effect of adding generated responses from LLMs to SQ, the differences between SQ alone and SQ combined with OsakaED (OsakaED+SQ), as well as between SQ alone and SQ combined with GPT-4 (GPT-4+SQ), were assessed. In the comparison between SQ and OsakaED+SQ, consistency was found to decrease ($p<0.05$). In the comparison between SQ and GPT-4+SQ, there were significant differences in user empathy, trust, agency, and emotion (p<0.05), as well as in mood, personality, and the system's empathy ($p<0.01$). When SQ was added to LLM-generated responses, the decrease in consistency remained notable in the case of OsakaED+SQ, suggesting that the addition of SQ does not necessarily mitigate the variability introduced by LLM responses. 
The findings from GPT-4+SQ comparisons reinforce the idea that GPT-4's responses significantly influence user experience, potentially overshadowing the impact of the added SQ.

\subsection{Comparison between generated responses and generated responses with SQ}
To investigate the effect of adding SQ to generated responses from LLMs, the differences between OsakaED and OsakaED+SQ, as well as between GPT-4 and GPT-4+SQ, were assessed.
However, no significant differences were found at any point. 
Surprisingly, contrary to our expectations, the addition of SQ to LLM-generated responses did not yield significant differences in any assessed dimension. 
This could indicate that the presence of SQ does not substantially alter the user's experience once they are engaged in LLM-generated content. 
This may be because the LLM is capable of generating a sufficiently informative or attractive response that including SQ would be redundant, but the situation would be different if SQ is added to the OsakaED than if it is added to GPT-4.
In the comparison between OsakaED and OsakaED+SQ, it is observed that 13 of 17 items were increased by adding SQ, including cognitive change, naturalness, and ease, although no significant differences were found.
On the other hand, in the comparison between GPT-4 and GPT-4+SQ, 14 items except for mood change were decreased by adding SQ.
This could possibly indicate that the presence of SQ improves the response of OsakaED but worsens the response of GPT-4. 
In particular,  it is noteworthy that both GPT-4 and GPT-4+SQ were observed to generate Socratic Questions. 
This observation suggests that the combination with SQ becomes more redundant, as GPT-4 alone already appears capable of producing such responses.

\section{Discussion}

\subsection{Principal Findings}
In this study, we constructed dialogue systems based on a CBT scenario focused on Socratic questioning using two kinds of LLMs. 
By comparing systems that used LLM-generated responses with those that did not, we investigated the impact of the generated responses. 
OsakaED does not necessarily improve outcomes compared to scenario-based dialogue, even though the dialogue model used to generate responses was trained with a real counseling dataset. 
On the other hand, the results suggest that GPT-4 is highly capable for counseling.
In particular, it's notable that the outcome and many of the dialogue qualities were often significantly improved when using GPT-4 generated responses, even though the user had just a six-turn interaction and the provided prompt (system message) for this was not so highly optimized.
In addition, It was observed that GPT-4 can ask Socratic Questioning in context.
This may provide many implications for the field in analyzing what features of the GPT-4 generated responses achieve these improvements in mood and response qualities. 

The fact that OsakaED did not show any particular improvement over SQ, a simple scenario-based system, indicates that building and using LLMs with a dataset of human counselors does not necessarily lead to an improvement in the system.
Therefore, when developing a dialogue system that uses LLMs to generate responses, it would be important to check beforehand whether the system can be improved compared to older systems such as rules, scenarios, and example responses.

On the other hand, The number of parameters for OsakaED was 1.6B, which was relatively small.
In recent years, open-source LLMs with a larger number of parameters, such as LLaMA \cite{touvron2023llama} and LLaMA 2 \cite{touvron2023llama2} by Meta, have been released. 
As there are already studies that have tuned these in the mental health field \cite{liu2023chatcounselor, yang2023mentalllama}, by producing a model aligned to CBT, it is thought to be possible to develop an LLM that surpasses the scenario-based dialogue systems in this setting.

\subsection{Ethical use of LLMs for response generation in mental healthcare services}
While GPT-4 was found to be able to generate high-quality responses, using GPT-4 in real-world mental health care services and interacting directly with users will raise ethical issues \cite{lee2023benefits, jeyaraman2023chatgpt, amamou2023chatgpt, gilbert2023large}.
First of all, even though mental health care services will be used by people with mental health issues, especially some of whom may have suicidal ideation, developers cannot control in advance what responses GPT-4 will generate and lead those users.
In other words, developers cannot fully guarantee the safety of the system's interaction with users, and this can be a critical ethical issue.
This includes the problem that LLMs may make up false information due to the ``hallucination" phenomenon they cause.
The issue of ``hallucination" is a critical shortcoming in this field because it can fabricate wrong information about medical care \cite{lee2023benefits, darcy2023generative}.
Although research on LLM alignment has been heavily studied, and alignment has been done for GPT-4 \cite{ouyang2022training}, it is unlikely that this fundamental concern of not being able to confirm output in advance will ever be resolved.
Actually, Woebot Health has a policy of outputting responses designed by human experts in advance, without direct user interaction with the LLMs \cite{darcy2023generative, woebot2023aiatwoebot}.

Given the ethical issues, it is reasonable to avoid directly presenting the response output by the LLMs to the user and to output a pre-designed response, as in the case of woebot.
Nevertheless, one way to take advantage of the LLMs high response generation capability is for a human expert to work with LLMs in advance to generate a dictionary of example responses in advance.
In this method, a large number of example responses for various contexts can be generated in advance using LLMs, and checked and supervised by human experts to generate some of the responses the systems chose from.
If there was no more appropriate response for the context, a human-in-the-loop strategy in which the human expert efficiently adds new types of responses using GPT-4 or other LLMs would allow more appropriate responses to be uttered at each turn that uses this response strategy.
In addition, the context and its response pairs produced during this cycle could be used to train other open-source LLMs, like LLaMA, to make better response generation models by instruction tuning, which will potentially accelerate these cycles \cite{peng2023instruction}.

Furthermore, in the construction of dialogue systems for mental health care, understanding user input through LLMs may also be an important application, which may be used to perform response selection and to estimate and monitor user states.
Recently, there have been studies on the estimation of various mental states of users focusing on depression, suicidal ideation, cognitive distortions, etc., using GPT-4 and other LLMs, and there is a possibility that the mental states of users can be understood more precisely and from various perspectives \cite{xu2023leveraging, yang2023mentalllama, levkovich2023suicide, chen2023empowering}.
It is expected that a real-world interactive system for mental health care will be constructed and operated as a human-in-the-loop operation, in which human experts provide supervision and intervention to improve the system while reaping the benefits of these high capabilities of LLMs.



\subsection{Limitations}
As a limitation, the experiment in this study is a small-scale one with a limited number of crowdsourced subjects. 
In particular, CBT is a long-term, multi-session program, and if its effectiveness as CBT is to be accurately verified, it is necessary to conduct a long-term randomized controlled trial with a larger sample size to compare its effectiveness.


Furthermore, in this experiment, the system was set up so that responses with and without LLMs were not returned at the same speed, but as quickly as possible from time to time.
As a result, responses without LLMs presented the next response instantaneously, while those with LLMs had a delay of several seconds and especially longer delays when GPT-4 was used.
It is known that the speed of response changes the user's evaluation of the system \cite{moon1999effects, holtgraves2007procedure, holtgraves2007perceiving, gnewuch2018faster, folstad2020users}, and it is possible that these responses speeds influenced this human evaluation.
In fact, some answers for free text questions asking for improvements to the system indicated that the responses were too fast, which led to comments such as, ``When I received a reply the moment I sent my message, I felt that I was getting a fixed reply without considering what I had written."
For the GPT-4 and GPT-4+SQ, comments were seen that responses were too slow, leading to comments such as, ``I was worried that the system had stopped."
In order to more accurately compare only the quality of the responses, it would be necessary to match the response speeds for all utterances.

\section{Conclusion}
In this study, we evaluated dialogue systems based on CBT scenarios using two types of LLMs: OsakaED and GPT-4. 
We compared systems with LLM-generated responses to those without, to assess the impact of these responses. 
The results suggest that GPT-4 is highly effective in counseling contexts, but it is also observed that OsakaED does not necessarily yield better outcomes than scenario-based dialogues, even though LLMs are being trained on real counseling data.
This could suggest that the use of LLMs is not always effective, compared to the scenario-based systems.
In addition, although GPT-4 is considered to be effective for mental healthcare response generation, it will raise safety problems if a system directly shows responses generated from LLMs to the users.
One possible approach to solving safety issues while taking advantage of the high response generation capability of LLMs such as GPT-4 is thought to develop a human-in-the-loop system that uses a dictionary of example responses or templates created in cooperation between LLMs and human experts.
In the future, we will focus on counseling by GPT-4, investigate optimally effective prompts for CBT, and analyze the needed counseling skills. Furthermore, we will verify the achieved counseling skills by the prompts.

%


\begin{acknowledgement}
This research is supported by CREST (grant number: JPMJCR19A5). The OsakaED dataset was provided by the Public Health and Medical Administration Office, Department of Public Health and Medical Affairs, Osaka Prefectural Government, and the Kansai Counseling Center.
\end{acknowledgement}

\bibliographystyle{plain}
\bibliography{ref}

\end{document}